\documentclass[conference]{IEEEtran}
\IEEEoverridecommandlockouts
\usepackage{booktabs}
\usepackage{makecell}
\usepackage{tabularray}
\usepackage{cite}
\usepackage{amsmath,amssymb,amsfonts}
\usepackage{algorithmicx}
\usepackage{graphicx}
\usepackage{textcomp}
\usepackage{xcolor} 
\def\BibTeX{{\rm B\kern-.05em{\sc i\kern-.025em b}\kern-.08em
    T\kern-.1667em\lower.7ex\hbox{E}\kern-.125emX}}
\DeclareUnicodeCharacter{FF1A}{:}
\DeclareUnicodeCharacter{00B7}{\textemdash}
\DeclareUnicodeCharacter{FF08}{(}
\DeclareUnicodeCharacter{FF09}{)}
\usepackage{algorithm}
\usepackage{algpseudocode}

\begin{document}

\title{Multi-tool Integration Application for Math Reasoning Using Large Language Model }

\author{\IEEEauthorblockN{1\textsuperscript{st} Zhihua Duan}
\IEEEauthorblockA{\textit{Intelligent Cloud Network Monitoring Department} \\
\textit{China Telecom Shanghai Company}\\
\textit{700 Daning Road, Shanghai, 200072}\\
Shanghai,China \\
duanzh.sh@chinatelecom.cn}
\and
\IEEEauthorblockN{2\textsuperscript{nd} Jialin Wang}
\IEEEauthorblockA{\textit{Computer Science} \\
\textit{Stanford University}\\
\textit{450 Serra Mall, Palo Alto,94305}\\
California, America \\
jialinwangspace@gmail.com}
 
}

\maketitle

\begin{abstract}
Mathematical reasoning is an important research direction in the field of artificial intelligence. This article proposes a novel multi tool application framework for mathematical reasoning, aiming to achieve more comprehensive and accurate mathematical reasoning by utilizing the collaborative effect of large language models (LLMs) and multiple external tools. Firstly, use a Math Tool to perform basic mathematical calculations during the inference process through interaction with LLM. Secondly, Code Tool can generate code fragments that comply with syntax rules and execute them, providing support for complex mathematical problems. Then, through the iterative reasoning of the CoT Tool, the logical coherence and accuracy of mathematical reasoning are enhanced. Ultimately, by using self consistency tools to select the final answer based on different parameters, the consistency and reliability of reasoning are improved. Through the synergistic effect of these tools, the framework has achieved significant performance improvement in mathematical reasoning tasks. We conducted experiments on the NumGLUE Task 4 test set, which includes 220 mathematical reasoning fill in the blank questions. The experimental results showed that, based on Math Tool, Code Tool, and CoT Tool, in Task 4 task,our method achieved an accuracy of 89.09,compared with the GPT3+FewShot baseline, Few Shot+ERNIE-4.0+self consistency improved by 49.09\%, and compared with fine-tuning the Fine tuning baseline, Few Shot+ERNIE-4.0+self consistency improved by 52.29\% 
\end{abstract}

\begin{IEEEkeywords}
 ERNIE-4.0,FewShot,CoT,Large Language Model
\end{IEEEkeywords}

\section{Introduction}
Mathematical reasoning is an important field in artificial intelligence research, which solves complex mathematical problems through deduction and reasoning under the guidance of logic and mathematical rules. However, for computers, conducting mathematical reasoning remains a challenging task. In recent years, with the rapid development of large language models, utilizing their powerful language generation and comprehension abilities to assist mathematical reasoning has become a new research direction.

Recent research has focused on improving the mathematical reasoning ability of Large Language Models (LLMs). By introducing Chain of Thinking (CoT) prompts, LLM has made progress in mathematical reasoning tasks. CoT prompts guide LLM to gradually solve problems, improving the accuracy and interpretability of reasoning. However, there are still some problems and limitations when dealing with scenarios such as common sense reasoning, formal logic, and algebraic computation. Current research is still focused on simple arithmetic reasoning, and for more complex mathematical concepts and problems, Further research is needed to expand the scope and ability of mathematical reasoning.
This article aims to propose a novel multi tool application framework for mathematical reasoning, utilizing a large language model driven approach and combining the collaborative effects of multiple external tools to achieve more comprehensive and accurate mathematical reasoning.

As shown in Figure 1, our framework utilizes various external tools such as Math Tool, Code Tool, CoT Tool, and self consistency tools in the inference process through a large language model to provide diverse inference support.

\begin{figure*}[htbp]
  \centering
  \includegraphics[width=\linewidth]{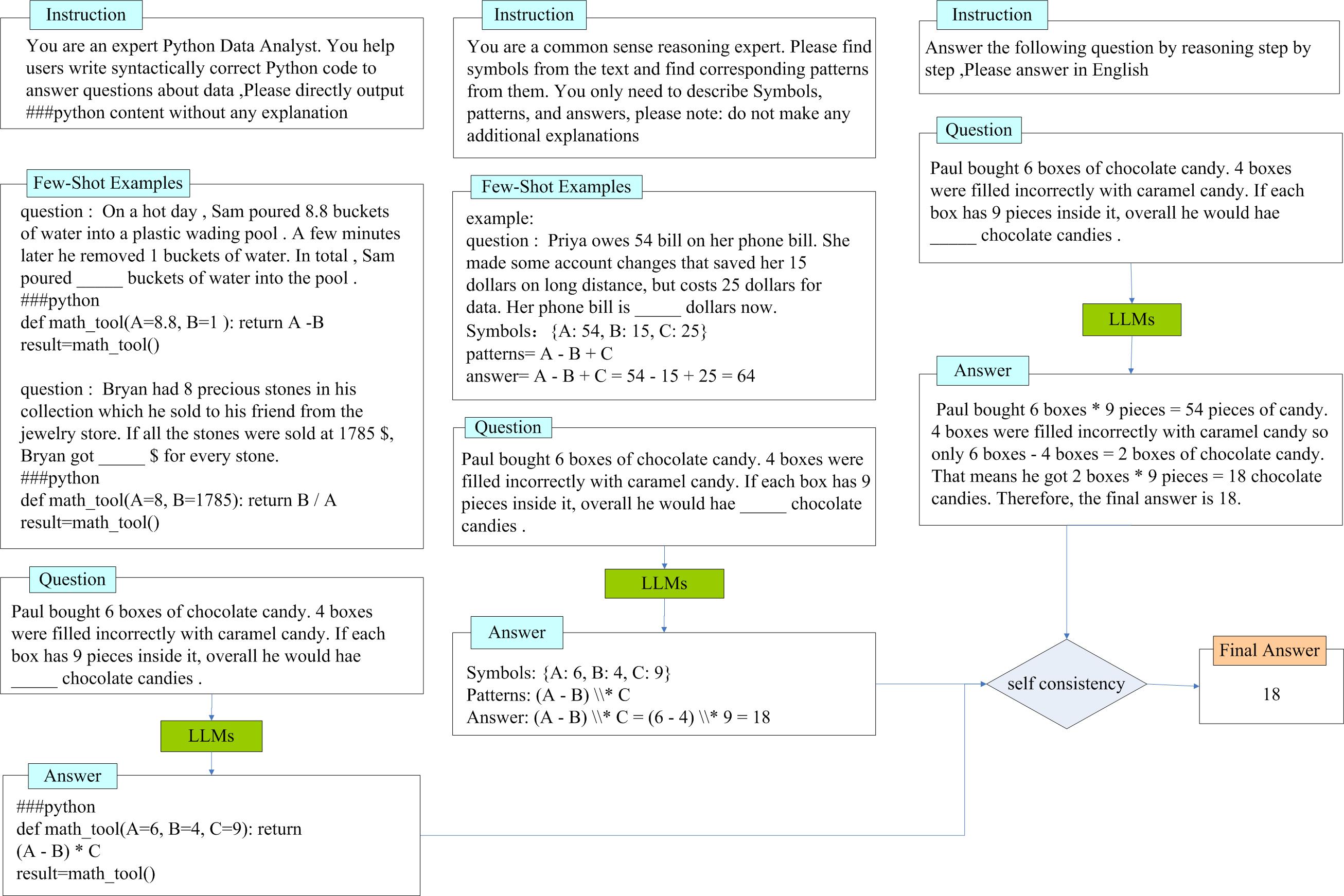}
  \caption{ Code Tool, Math Tool, CoT Tool, Self Consistency Tool Architecture Diagram}
   
\end{figure*}

The unique contribution of this paper lies in the implementation of a self-consistency tool. As shown in Figure 2, based on the parameter configuration, the mathematical calculator, code executor, and thought chain tool are sequentially selected to obtain answers. If all three tools are used simultaneously, the answer with the highest occurrence count is chosen as the final answer. If each answer appears only once, the answer from the code is given priority based on the configured priority.

\section{Related Work}
In mathematical reasoning tasks, the MultiTool CoT framework combines multiple external tools such as calculators and knowledge retrievers, significantly improving the performance of large language models in digital reasoning tasks\cite{multi-tools}. MathPrompt technology improves the performance of large language models on arithmetic problems by generating multiple algebraic expressions or Python functions to solve the same mathematical problem \cite{MathPrompter}. The use of prompt based learning paradigms can improve the performance of information extraction tasks. CodeIE proposes a method to convert structured output into code form and uses a code generation language model to perform named entity recognition and relationship extraction tasks \cite{CodeIE}. NumGLUE is a multitasking benchmark used to evaluate the performance of artificial intelligence systems on eight different tasks, promoting cross task knowledge sharing \cite{NumGlue}. MathWorld is a graph based semantic formalism specifically used in the field of mathematical story problems. By using MathWorld, the world model can be associated with mathematical story problems, representing the context, actions, and mathematical relationships introduced in the text \cite{world_math}. LogicSolver first retrieves highly relevant algebraic knowledge for each mathematical text problem, and then passes them as prompts to the backbone model to improve the semantic representation of the mathematical text problem \cite{LogicSolver}. MAmmoTH is a large-scale language model specifically designed for solving general mathematical problems, emphasizing the importance of diverse problem coverage\cite{MAmmoTH}. In complex mathematical related tasks, a step-by-step reasoning approach is used to initialize the solution through retrieved samples, and then the intermediate steps of the generated solution are checked and refined from the perspectives of tool operation and natural language reasoning until a convergent solution is obtained.

In contrast to the preceding efforts, this study introduces an innovative methodology within the domain of mathematical reasoning that synergistically integrates the capabilities of large language models alongside various auxiliary tools such as Math Tool, Code Tool, and CoT Tool, all designed to augment the capacity for mathematical reasoning.

\section{Methods}
The mathematical reasoning multi tool application we propose is an interactive framework that allows LLM to use multiple external tools during the reasoning process: Math Tool, Code Tool, Cot Tool, and self consistency Tool.

In Math Tool, the symbols used in prompts have little impact on model performance, which may be counterintuitive, but patterns as a means of enhancing task understanding\cite{text-pattern} will prompt the model to generate correct output. Most importantly, text and patterns form a symbiotic relationship and play an important role in mathematical reasoning. Text helps generate useful patterns, The Math Tool is shown in Table I. such as extracting mathematical patterns, which enhance task understanding and enable language models to generate text that helps solve tasks. The success of Math Tool is attributed to the interaction between text and patterns, applying extracted symbols to mathematical patterns. This is of great significance for further improving and optimizing the application of large language models.

\begin{table}
\centering
\caption{MATH TOOL}
\begin{tblr}{
  width = \linewidth,
  colspec = {Q[40]Q[896]},
  hlines,
}
Q: & A construction company bought 8.11 tons of sand and sold 5.91 tons of gravel.In total , the  company has \_\_\_\_\_ tons of material. \\
A: & {Symbols：\{A: 8.11, B: 5.91\}
\\patterns =  A - B 
\\answer = A - B = 8.11 -5.91 = 2.2 }    
\end{tblr}
\end{table}

Code Tool is a Python code execution function, as shown in Table II. Its main function is to call Baidu Big Model Service to generate code snippets that comply with syntax rules based on user input prompts. The tool first retrieves Python function text by calling Baidu's Big Model service, and dynamically executes the code using the built-in function exec(). The exec() function is capable of executing complex Python statements, receiving Python code stored in strings or objects, and returning the processed answer, which is the result of the function execution.

\begin{table}
\centering
\caption{code tool}
\begin{tblr}{
  width = \linewidth,
  colspec = {Q[40]Q[896]},
  hlines,
}
Q: & The Richmond Tigers sold ticket last season.  They sold 9570 tickets at the gate and then an additional 3867 tickets online. \_\_\_\_\_ tickets were sold in total. \\
A: & {\#\#\# python
\\def math\_tool(A = 9570, B = 3867) :  return A + B
\\result = math\_tool()}                                                                        
\end{tblr}
\end{table}

The CoT Tool,as shown in Table III. Its function is to infer based on the input thinking chain prompt words by calling Baidu's big model service to obtain the result of thinking chain inference. This tool uses iterative reasoning to gradually extract the final answer from the reasoning text by calling Baidu's big model service again. 

\begin{table}
\centering
\caption{CoT tool}
\begin{tblr}{
  width = \linewidth,
  colspec = {Q[40]Q[896]},
  hlines,
}
Q: & Paul bought 6 boxes of chocolate candy. 4 boxes were filled incorrectly with caramel candy. If each box has 9 pieces inside it, overall he would have \_\_\_\_\_ chocolate candies.                                                                                               \\
A: & {Paul bought 6 boxes * 9 pieces = 54 pieces of candy. 
\\4 boxes were filled incorrectly with caramel candy, 
\\so only 6 boxes - 4 boxes = 2 boxes of chocolate candy. 
\\That means he got 2 boxes * 9 pieces = 18 chocolate candies. 
\\Therefore, the final answer is 18.} 
\end{tblr}
\end{table}

The self consistency tool implements a decision system that selects different answers based on given parameters. If the self consistency feature is enabled, the system will call three different tools: Math Tool), Code Tool, and CoT Tool. Firstly, the system will call the Math Tool, Code Tool, and CoT Tool to obtain three answers respectively, and add these answers to a list. Then, the system will count the number of times each answer appears in the list and select the answer with the most occurrences as the final answer. If each answer only appears once, the answer with the highest priority will be selected as the final answer based on the pre-set priority.

\begin{figure}[htbp]
  \centering
  \includegraphics[width=\linewidth]{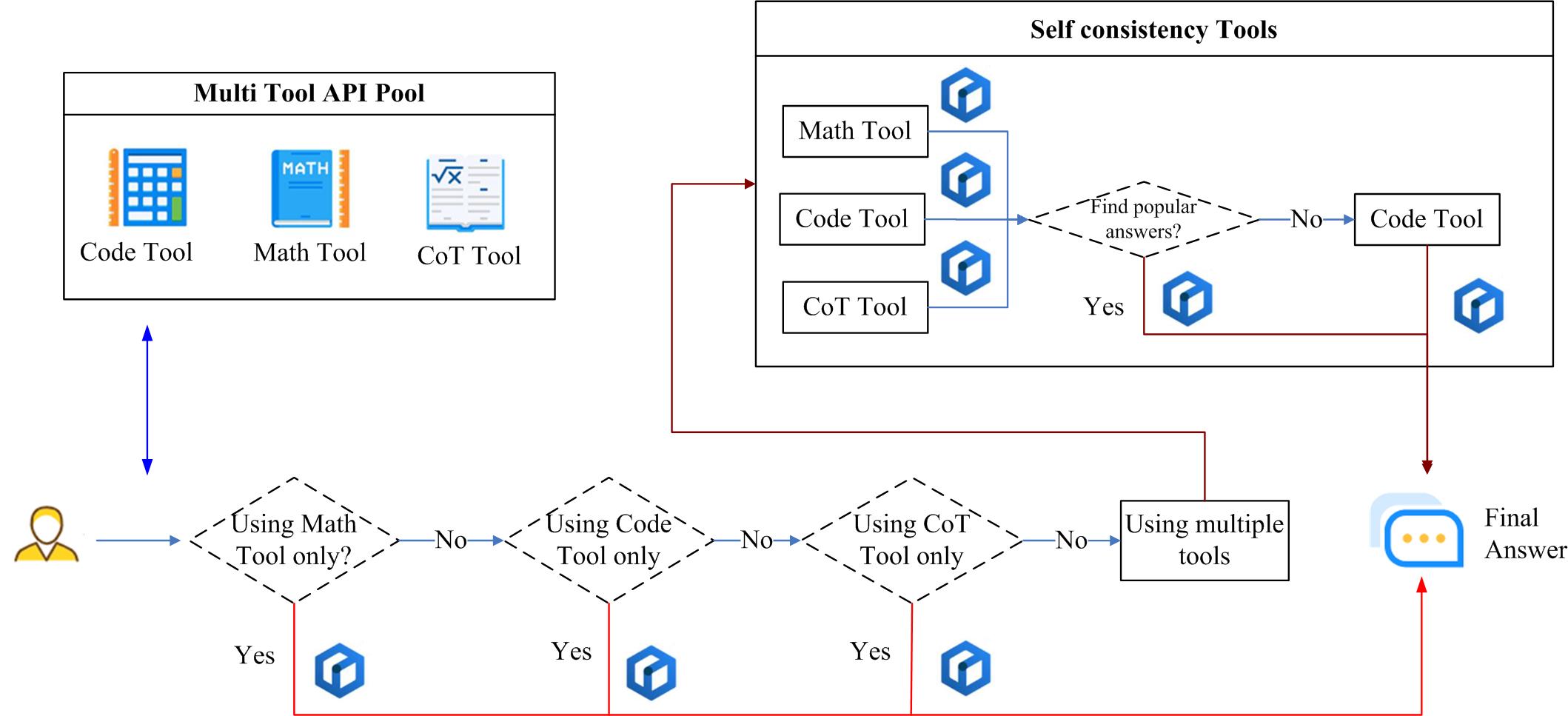}
  \caption{ Self consistent tool Diagram}
   
\end{figure}
 
\section{Experiment}

\subsection{DataSets}\label{subsec1}
NumGLUE is a multitasking dataset consisting of 8 different tasks. Task 1 is common sense+arithmetic, Task 2 is domain specific knowledge+arithmetic, Task 3 is common sense+quantitative, Task 4 is fill in the blank, Task 5 is reading comprehension+explicit numerical reasoning, Task 6 is reading comprehension+implicit numerical reasoning, Task 7 is quantitative natural language reasoning, and Task 8 is arithmetic problem. These tasks involve common sense, domain specific knowledge, and quantitative reasoning Different aspects such as fill in the blank questions and reading comprehension. Through this dataset, the performance of different models on various tasks can be evaluated.

As shown in Table IV, Task 4 is a Fill in the blank dataset, which retrieves questions from an arithmetic question bank \cite{NumGlueArithmetic} \cite{NumGlueUnit} \cite{NumGlue_Knowledge},and converts them into the format of fill in the blank questions. Require the generation of correct fill in the blank answers based on the given context, and provide understanding and answers to mathematical problems through fill in the blank questions. 

This dataset consists of three parts: training set, validation set , and test set. There are 770 samples in the training set, 110 samples in the validation set, and 220 samples in the test set. This article uses Few Shot+LLms to directly test 220 samples from the test set.

\begin{table}
\centering
\caption{Task4 Fill-in-the-blanks DataSets}
\begin{tblr}{
  width = \linewidth,
  colspec = {Q[75]Q[46]Q[812]},
  hlines,
}
Dataset & Size & Example                                                                                                                                                              \\
train   & 770  & {Question: Joan found 70 seashells on the beach. She gave Sam some of her seashells. She has 27 seasshells left. She gave \_\_\_\_\_ seashells to Sam? Answer: 43} \\
dev     & 110  & Sally had 760 quarters in her bank. She received 418 more quarters. She has \_\_\_\_\_ quarters now? Answer: 1178                                                    \\
test    & 220  & If I buy 80 gumdrops and each gumdrop costs 4 cents, So , I need \_\_\_\_\_ cents.Answer: 320                                                                        
\end{tblr}
\end{table} 

\subsection{Method comparison}\label{subsec1}

As shown in Table V, ERNIE-4.0 has achieved good performance through self consistency constraints. Under the self consistency constraint, if the results of each tool appear the same number of times, the reasoning answer of the thought chain CoT is prioritized, and the performance of ERNIE-4.0 reaches 75.9. When obtaining the results of the generated code function first, the performance further improves to 80.45. In the case of ERNIE-4.0, by combining Math Tool, CoT Tool, and Code Tool to prioritize obtaining the results of Math Tool, the performance of ERNIE-4.0 reached an impressive 89.09. Compared to the GPT3+FewShot baseline, ERNIE-4.0 improved by 49.09\% (=89.09-40). Compared to fine-tuning the Fine tuning baseline, ERNIE-4.0 improved by 52.29\% (=89.09-36.8)

\begin{table}[h]
\centering
\caption{Performance in the Task 4 dataset of NumGLUE.Scores with $^{\dag}$~  were ~  obtained ~ from ~\cite{NumGlue}.}
\begin{tblr}{
  width = \linewidth,
  colspec = {Q[173]Q[494]Q[185]Q[77]},
  hline{1-2,4,6,10-11,18,27-28} = {-}{},
}
Learning    & Baseline Category                          & Baseline Name & Task4 \\
HEURISTIC$^{\dag}$   & Task-specific                              & Random        & 0     \\
            & Task-specific                              & Majority      & 0.5   \\
ZERO-SHOT$^{\dag}$   & -                                          & GPT3          & 2     \\
            & -                                          & GPT3-Instruct & 3     \\
FEW-SHOT$^{\dag}$    & Task-specific                              & GPT3          & 40    \\
            & Task-specific                              & GPT3-Instruct & 33    \\
            & Multi-task                                 & GPT3          & 1     \\
            & Multi-task                                 & GPT3-Instruct & 2     \\
FINE-TUNING$^{\dag}$ & Multi-task                                 & GPT3-13B      & 11.1  \\
FINE-TUNING$^{\dag}$ & Multi-task(Q-only)                         & Ex-NumNet     & 0.5   \\
            & Multi-task(C-only)                         & Ex-NumNet     & 19.1  \\
            & Single-task                                & Ex-NumNet     & 22.2  \\
            & Multi-task                                 & Ex-NumNet     & 31.4  \\
            & Multi-task+IR                              & Ex-NumNet     & 36.4  \\
            & Multi-task+CIR                             & Ex-NumNet     & 36.8  \\
            & Multi-task+OS                              & Ex-NumNet     & 35.9  \\
MultiTool   & Zero-Shot + CoT                            & ERNIE-4.0     & 64.09 \\
            & Few-Shot(1 samples,math tool only)         & ERNIE-4.0     & 60.00    \\
            & Few-Shot(5 samples,math tool only)         & ERNIE-4.0     & 60.00    \\
            & Few-Shot(1 samples,code tool only)         & ERNIE-4.0     & 74.09 \\
            & Few-Shot(5 samples,code tool only)         & ERNIE-4.0     & 77.72 \\
            & Few-Shot + self consistency                & ERNIE-4.0     & 71.81 \\
            & Few-Shot + self consistency ( CoT prior )  & ERNIE-4.0     & 75.90  \\
            & Few-Shot + self consistency ( Code prior ) & ERNIE-4.0     & 80.45 \\
            & Few-Shot + self consistency ( Math prior ) & ERNIE-4.0     & 89.09 \\
            &                                            & Human         & 95    
\end{tblr} 
\end{table}
  
\section{CONCLUSION}
This study successfully implemented a multi tool application framework for mathematical reasoning based on a large language model, which utilizes multiple external tools during the reasoning process, including Math Tool, Code Tool, and CoT Tool. Math Tool can perform basic mathematical calculations, code executor tools can generate code fragments that conform to syntax rules and execute them, and CoT Tool obtain the results of the inference chain through iterative reasoning. The synergistic effect of these external tools has enabled our framework to perform well in mathematical reasoning tasks. The design of this framework is universal and can be applied to various tasks by extending more external tools. Future work can further explore and optimize the selection and integration of external tools within the framework to improve inference efficiency and performance, and apply the framework to a wider range of fields and practical scenarios.   
 
\section*{ACKNOWLEDGMENT}
This research was sponsored by Wenxin Model 4.0, a large model platform of China Baidu AI Cloud Qianfan. ERNIE-4.0 is a large language model independently developed by Baidu, covering a massive amount of Chinese data and possessing stronger abilities in dialogue, question answering, and content creation.

\bibliographystyle{unsrt}

\end{document}